\documentclass{llncs}
\usepackage{llncsdoc}
\usepackage{url}
\usepackage[numbers]{natbib}
\usepackage{times}
\usepackage{latexsym}
\usepackage{amsmath}
\usepackage{dsfont}
\usepackage[subrefformat=parens,labelformat=parens]{subfig}
\usepackage{graphicx}
\usepackage{color}
\usepackage[normalem]{ulem}
\usepackage{multirow}
\usepackage{amssymb}
\usepackage{amsfonts}
\usepackage{times} 

\newcommand{\smlmath}[1]{\mbox{\small $#1$}}
\begin{document}
\title{Large-scale Multi-label Text Classification --- \\ Revisiting Neural Networks}

\author{Jinseok Nam\inst{1,2} \and Jungi Kim\inst{1} \and Eneldo Loza Menc{\'i}a\inst{1} \and\\
Iryna Gurevych\inst{1,2} \and Johannes F{\"u}rnkranz\inst{1}}

\institute{
Department of Computer Science, Technische Universit\"{a}t Darmstadt, Germany
\and
Information Center for Education, German Institute for Educational Research, Germany
}

\maketitle

\begin{abstract}
  Neural networks have recently been proposed for 
  multi-label classification because they are able to capture and
  model label dependencies in the output layer.  In this work, we
  investigate limitations of BP-MLL, a neural network (NN)
  architecture that aims at minimizing pairwise ranking error.
  Instead, we propose to use a comparably simple NN approach 
  with recently proposed learning techniques for large-scale multi-label
  text classification tasks.  In particular, we show that BP-MLL's ranking
  loss minimization can be efficiently and effectively replaced with
  the commonly used cross entropy error function, and demonstrate that
  several advances in neural network training that have been 
  developed in the realm of deep learning can be effectively employed
  in this setting.
  Our experimental results show that
  simple NN models equipped with 
  advanced techniques such as
  rectified linear units, dropout, and AdaGrad perform as well as or
  even outperform state-of-the-art approaches on six large-scale
  textual datasets with diverse characteristics.
\end{abstract}

\section{Introduction} \label{SEC_Introduction}

As the amount of textual data on the web and in digital libraries is increasing rapidly, the need for augmenting unstructured data with metadata is also increasing.
Systematically maintaining a high quality digital library requires extracting a variety types of information from unstructured text, from trivial information such as title and author, to non-trivial information such as descriptive keywords and categories.
Time- and cost-wise, a manual extraction of such information from ever-growing document collections is impractical.

Multi-label classification is an automatic approach for addressing such problems by learning to assign a suitable subset of categories from an established classification system to a given text.
In the literature, one can find a number of multi-label classification approaches for a variety of tasks in different domains such as bioinformatics \cite{bi2011multi}, music \cite{tsoumakas2008multi}, and text \cite{Fuernkranz2008}.
In the simplest case, multi-label classification may be viewed as a set of binary classification tasks that decides for each label independently whether it should be assigned to the document or not.
However, this so-called \emph{binary relevance} approach ignores dependencies between the labels, so that current research in multi-label classification concentrates on the question of how such dependencies can be exploited \cite{Read2011,Dembczy2010}.
%
%
%
One such approach is BP-MLL \cite{Zhang2006}, which formulates
multi-label classification problems as a neural network with multiple
output nodes, one for each label. The output layer is able to
model dependencies between the individual labels.

In this work, we directly build upon BP-MLL and show how a simple,
single hidden layer NN may achieve a state-of-the-art performance in
large-scale multi-label text classification tasks. The key
modifications that we suggest are (i) more efficient and more effective
training by replacing BP-MLL's pairwise ranking loss with cross
entropy
and (ii) the use of recent developments in the area of deep learning such as rectified linear
units (ReLUs), Dropout, and AdaGrad.

Even though we employ techniques that have been developed in the realm
of deep learning, we nevertheless stick to single-layer NNs.
The motivation behind this is two-fold: first, a simple network configuration allows better scalability of the model and is more suitable for large-scale tasks.
Second, as it has been shown in the literature \cite{Joachims1998}, popular feature representation schemes for textual data such as variants of tf-idf term weighting already incorporate a certain degree of higher dimensional features, and we speculate that even a single-layer NN model can work well with text data.
This paper provides an empirical evidence to support that
a simple NN model equipped with recent advanced techniques for
training NN performs as well as or even outperforms state-of-the-art
approaches on large-scale datasets with diverse characteristics.

\section{Multi-label Classification}

Formally, multi-label classification may be defined as follows: $X
\subset \mathbb{R}^D$ is
a set of $M$ instances, each being a $D$-dimensional feature vector, and
$L$ is a set of labels. Each instance $\mathbf{x}$ is associated with
a subset of the $L$ labels, the so-called \emph{relevant} labels; all
other labels are \emph{irrelevant} for this example. The task of the
learner is to learn a mapping function $f:\mathbb{R}^{D} \rightarrow 2^L$ that
assigns a subset of labels to a given instance. An alternative view is
that we have to predict an $L$-dimensional target vector
$\mathbf{y}\in\{0,1\}^{L}$, where $y_i = 1$ indicates that the $i$-th
label is relevant, whereas $y_i = 0$ indicates that it is irrelevant
for the given instance.

Many algorithms have been developed for tackling this type of problem.
The most straightforward way is binary relevance (BR) learning;
it constructs $L$ binary classifiers, which are trained on the $L$
labels independently. Thus, the prediction of the label set is
composed of independent predictions for individual labels.
However, labels often occur together, that is, the presence of a
specific label may suppress or exhibit the likelihood of other labels.

To address this limitation of BR, pairwise decomposition (PW) and
label powerset (LP) approaches consider label dependencies during the
transformation by either generating pairwise subproblems
\cite{Fuernkranz2008,Mencia2010} or the powerset of possible label combinations
\cite{Tsoumakas2011}. Classifier chains \cite{Read2011,Dembczy2010}
are another popular approach that extend BR by including previous
predictions into the predictions of subsequent labels.



\cite{Elisseeff2001} present a large-margin classifier, RankSVM, that
minimizes a ranking loss by penalizing incorrectly ordered pairs of
labels. This setting can be used for multi-label classification by
assuming that the ranking algorithm has to rank each relevant label
before each irrelevant label. In order to make a prediction, the
ranking has to be \emph{calibrated} \cite{Fuernkranz2008}, i.e., a
threshold has to be found that splits the ranking into relevant and
irrelevant labels.
Similarly, \citet{Zhang2006} introduced a framework that learns ranking errors in neural networks via backpropagation (BP-MLL).


\subsection{State-of-the-art multi-label classifiers and limitations}
\label{SEC_PrevApproach}

The most prominent learning method for multi-label text classification
is to use a BR approach with strong binary classifiers such as SVMs \cite{Rubin2012,Yang2012} despite its simplicity.
It is well known that characteristics of high-dimensional and sparse
data, such as text data, make decision problems linearly separable
\cite{Joachims1998}, and this characteristic suits the strengths of
SVM classifiers well.

Unlike benchmark datasets, real-world text collections consist of a large number of training examples represented in a high-dimensional space with a large amount of labels.
To handle such datasets, researchers have derived efficient
\textit{linear} SVMs \cite{Joachims2006,Fan2008} that can handle
large-scale problems.
The training time of these solvers scales linearly with the number of instances, so that they show good performance on standard benchmarks.
However, their performance decreases as the number of labels grows and
the label frequency distribution becomes skewed \cite{Liu2005,Rubin2012}.
In such cases, it is also intractable to employ methods that minimize ranking errors among labels \cite{Elisseeff2001,Zhang2006} or that learn joint probability distributions of labels \cite{Ghamrawi2005,Dembczy2010}.

\section{Neural Networks for Multi-label Classification}

\begin{figure}[t]
\centering
\subfloat [A neural network]{\label{fig:NN} \includegraphics[bb=0 0 246 251,scale=0.3]{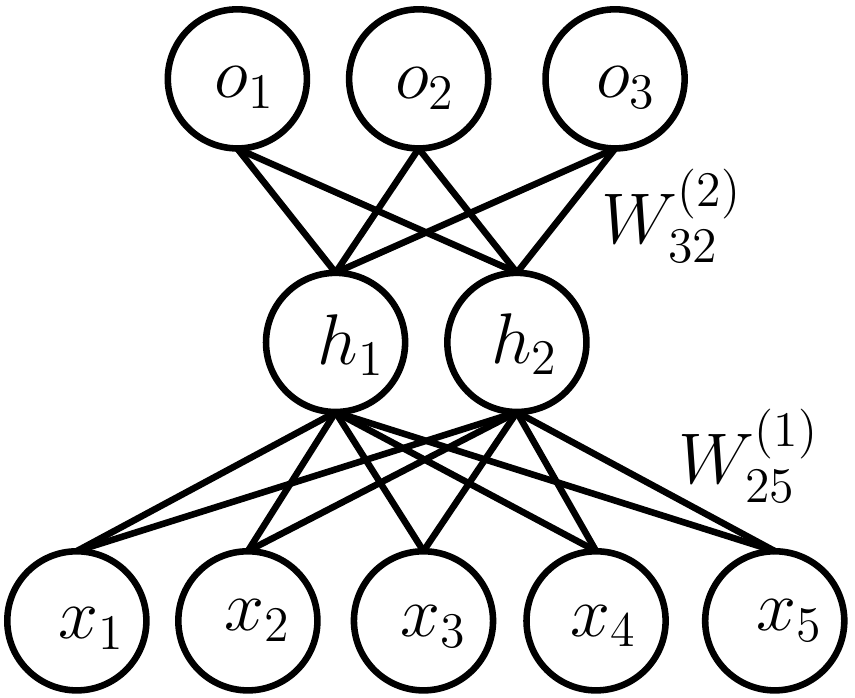}}
\qquad \qquad
\subfloat [Threshold decision]{\label{fig:Threshold} \includegraphics[bb=0 0 644 190,scale=0.3]{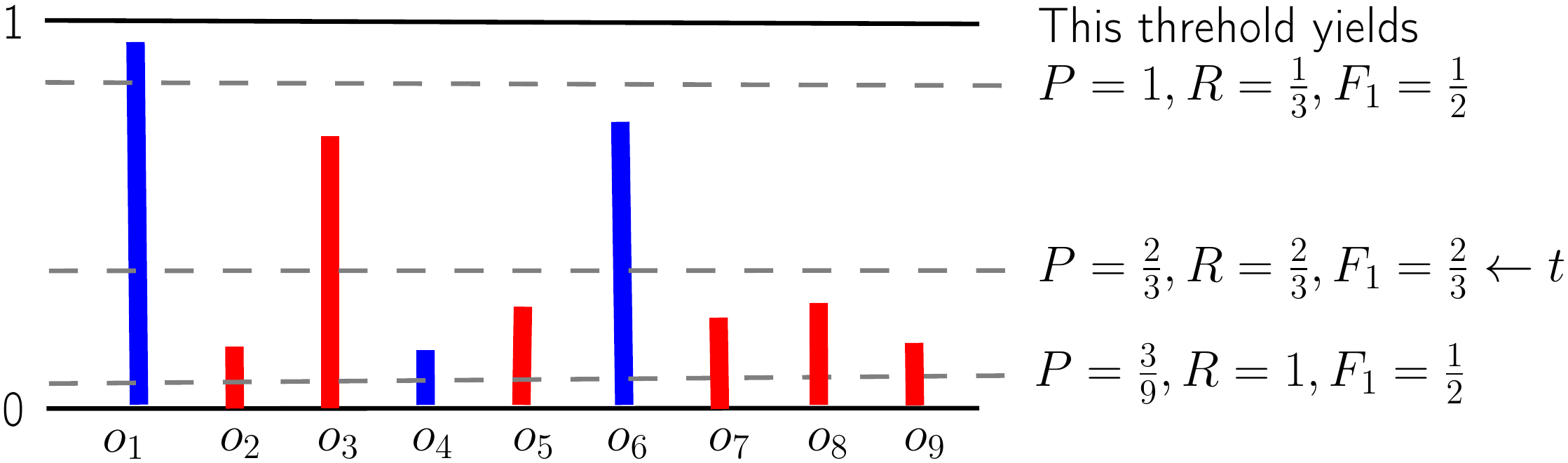}}
\caption{\protect\subref{fig:NN} a neural network with a single hidden
  layer of two units and multiple output units, one for each possible label.
\protect\subref{fig:Threshold} shows how threshold for a training
example is estimated based on prediction output $\mathbf{o}$ of the network.
Consider nine possible labels, of which $o_1$, $o_4$ and $o_6$
are relevant labels (blue) and the rest are irrelevant (red).
The figure shows three exemplary threshold candidates (dashed lines),
of which the middle one is the best choice because it gives the
highest F1 score. See Section \protect\ref{SEC_Thresholding} for more
details.}
\label{FIG_OVERALL_APPROACH}
\end{figure}

In this section, we propose a neural network-based multi-label
classification framework that is composed of a single hidden layer and
operates with recent developments in neural network and optimization
techniques, which allow the model to converge into good regions of the error surface in a few steps of parameter updates.
Our approach consists of two modules (Figure
\ref{FIG_OVERALL_APPROACH}): a neural network that produces label
scores (Sections~\ref{SEC_NN}--\ref{SEC_ADVANCES}), and a label predictor that converts label scores into binary
using a thresholding technique (Section \ref{SEC_Thresholding}).

\subsection{Rank Loss} \label{SEC_DefRankLoss}
The most intuitive objective for multi-label learning is to minimize the number of mis-ordering between a pair of relevant label and irrelevant label, which is called \textit{rank loss}:
\begin{small}
\begin{equation}
L(\mathbf{y},f(\mathbf{x}))=w(\mathbf{y})\sum_{y_{i}<y_{j}} \mathbb{I}\left(f_{i}(\mathbf{x}) > f_{j}(\mathbf{x})\right)+\frac{1}{2}\mathbb{I}\left(f_{i} = f_{j}\right)
\label{eq:rank_loss}
\end{equation}
\end{small}
where $w(\mathbf{y})$ is a normalization factor, $\mathbb{I}\left(\cdot\right)$ is the indicator function, and $f_{i}\left(\cdot\right)$ is a prediction score for a label $i$.
Unfortunately, it is hard to minimize due to non-convex property of the loss function.
Therefore, convex surrogate losses have been proposed as alternatives to rank loss \citep{Schapire2000,Elisseeff2001,Zhang2006}.

\subsection{Pairwise Ranking Loss Minimization in Neural Networks} \label{SEC_NN}
Let us assume that we would like to make a prediction on $L$ labels from $D$ dimensional input features. Consider the neural network model with a single hidden layer in which $F$ hidden units are defined and input units $\mathbf{x} \in \mathbb{R}^{D \times 1}$ are connected to hidden units $\mathbf{h} \in \mathbb{R}^{F \times 1}$ with weights $\mathbf{W}^{(1)} \in \mathbb{R}^{F \times D}$ and biases $\mathbf{b}^{(1)} \in \mathbb{R}^{F \times 1}$. The hidden units are connected to output units $\mathbf{o} \in \mathbb{R}^{L \times 1}$ through weights $\mathbf{W}^{(2)} \in \mathbb{R}^{L \times F}$ and biases $\mathbf{b}^{(2)} \in \mathbb{R}^{L \times 1}$.
The network, then, can be written in a matrix-vector form, and we can construct a feed-forward network $\smlmath{f_{\Theta}: \mathbf{x} \rightarrow \mathbf{o}}$ as a composite of non-linear functions in the range $[0,1]$:
\begin{small}
\begin{equation}
f_{\Theta}(\mathbf{x}) = f_{o}\left(\mathbf{W}^{(2)}f_{h}\left(\mathbf{W}^{(1)}\mathbf{x}+\mathbf{b}^{(1)}\right)+\mathbf{b}^{(2)}\right)
\label{eq:ff_func}
\end{equation}
\end{small}
\noindent
where $\smlmath{\Theta = \{\mathbf{W}^{(1)},\mathbf{b}^{(1)},\mathbf{W}^{(2)},\mathbf{b}^{(2)}\}}$, and $f_{o}$ and $f_{h}$ are \textit{element-wise} activation functions in the output layer and the hidden layer, respectively.
Specifically, the function $f_{\Theta}\left(\mathbf{x}\right)$ can be re-written as follows:
\begin{small}
\begin{align*}
\mathbf{z}^{\left(1\right)} &= \mathbf{W}^{\left(1\right)}\mathbf{x} + \mathbf{b}^{\left(1\right)}, \quad
\mathbf{h} = f_{h}\left(\mathbf{z}^{\left(1\right)}\right) \\
\mathbf{z}^{\left(2\right)} &= \mathbf{W}^{\left(2\right)}\mathbf{h} + \mathbf{b}^{\left(2\right)}, \quad
\mathbf{o} = f_{o}\left(\mathbf{z}^{\left(2\right)}\right)
\end{align*}
\end{small}
where $\mathbf{z}^{\left(1\right)}$ and $\mathbf{z}^{\left(2\right)}$ denote the weighted sum of inputs and hidden activations, respectively.

Our aim is to find a parameter vector $\smlmath{\Theta}$ that minimizes a cost function $J(\Theta;\mathbf{x}, \mathbf{y})$.
The cost function measures discrepancy between predictions of the
network and given targets $\mathbf{y}$.

BP-MLL \cite{Zhang2006} minimizes errors induced by incorrectly
ordered pairs of labels, in order to exploit dependencies among
labels. To this end, it introduces a \emph{pairwise error function}
(PWE), which is defined as follows:
\begin{small}
\begin{equation}
J_{PWE}(\Theta;\mathbf{x},\mathbf{y}) = \frac{1}{|\mathbf{y}||\bar{\mathbf{y}}|}\sum_{(p,n)\\ \in \mathbf{y}\times\bar{\mathbf{y}}} \exp(-(o_{p} - o_{n}))
\label{eq:err_BP-MLL}
\end{equation}
\end{small}
\noindent
where $p$ and $n$ are positive and negative label index associated with training example $\mathbf{x}$.
$\bar{\mathbf{y}}$ represents a set of negative labels and $|\cdot|$ stands for the cardinality.
The PWE is relaxation of the loss function in Equation \ref{eq:rank_loss} that we want to minimize.

As no closed-form solution exists to minimize the cost function, we use a gradient-based optimization method.
\begin{small}
\begin{equation}
\Theta^{(\tau+1)} = \Theta^{(\tau)} - \mathbf{\eta} \nabla_{{\Theta}^{(\tau)}}J(\Theta^{(\tau)};\mathbf{x},\mathbf{y})
\label{eq:grad_update}
\end{equation}
\end{small}
The parameter $\Theta$ is updated by adding a small step of negative gradients of the cost function $\smlmath{ J(\Theta^{(\tau)};\mathbf{x},\mathbf{y}) }$ with respect to the parameter $\Theta$ at step $\tau$.
The parameter $\mathbf{\eta}$, called the learning rate, determines the step size of updates.
\subsection{Thresholding}
\label{SEC_Thresholding}

Once training of the neural network is finished, its output may be
interpreted as a probability distribution
$p\left(\mathbf{o}|\mathbf{x}\right) $ over the labels for a given document $\mathbf{x}$.
The probability distribution can be used to rank labels, but
additional measures are needed in order to split the ranking into
relevant and irrelevant labels.
For transforming the ranked list of labels into a set of binary predictions, we train a multi-label threshold predictor from training data.
This sort of thresholding methods are also used in \citep{Elisseeff2001,Zhang2006}

For each document $\mathbf{x}_m$, labels are sorted by the probabilities in decreasing order. 
Ideally, if NNs successfully learn a mapping function $\smlmath{f_{\Theta}}$, all correct (positive) labels will be placed on top of the sorted list and there should be large margin between the set of positive labels and the set of negative labels.
Using $F_1$ score as a reference measure, we calculate classification performances at every pair of successive positive labels and choose a threshold value $t_{m}$ that produces the best performance (Figure~\ref{FIG_OVERALL_APPROACH}~\protect\subref{fig:Threshold}).

Afterwards, we can train a multi-label thresholding predictor $\hat{\mathbf{t}}=T\left(\mathbf{x};\theta\right)$ to learn $\mathbf{t}$ as target values from input pattern $\mathbf{x}$.
We use linear regression with $\ell$2-regularization to learn $\theta$
\begin{small}
\begin{equation}
J\left(\theta\right)=\frac{1}{2M}\sum_{m=1}^{M}\left(T\left(\mathbf{x}_{m};\theta\right) - t_{i}\right)^2+\frac{\lambda}{2}\|\theta\|_{2}^{2}
\label{eq:lin_reg}
\end{equation}
\end{small}
where $ T\left(\mathbf{x}_{m};\theta\right)=\theta^{T}\mathbf{x}_{m} $
and $\lambda$ is a parameter which controls the magnitude of the $\ell2$ penalty.

At test time, these learned thresholds are used to predict a binary output $\hat{y}_{kl}$ for label $l$ of a test document $\mathbf{x}_{k}$ given label probabilities $o_{kl}$; $\hat{y}_{kl}=1$ if $\smlmath{o_{kl} > T\left(\mathbf{x}_{k};\theta\right)}$, otherwise 0.

\subsection{Ranking Loss vs. Cross Entropy} \label{SEC_RankLoss}
BP-MLL is supposed to perform better in multi-label problems since it takes label correlations into consideration than the standard form of NN that does not.
However, we have found that BP-MLL does not perform as expected in our
preliminary experiments, particularly, on datasets in textual domain.

\paragraph*{\bf Consistency w.r.t Rank Loss} Recently, it has been claimed that none of convex loss functions including BP-MLL's loss function (Equation \ref{eq:err_BP-MLL}) is  consistent with respect to \textit{rank loss} which is non-convex and has discontinuity \cite{Clement2012,Gao2013}.
Furthermore, univariate surrogate loss functions such as \textit{log loss} are rather consistent with rank loss \cite{Dembczynski2012}.
\begin{small}
\begin{equation*}
J_{log}(\Theta;\mathbf{x},\mathbf{y}) = w\left(\mathbf{y}\right)\sum_{l}\log\left(1+e^{-\dot{y}_{l}z_{l}}\right)
\end{equation*}
\end{small}
where $w\left(\mathbf{y}\right)$ is a weighting function that normalizes loss in terms of $\mathbf{y}$ and $z_l$ indicates prediction for label $l$. Please note that the log loss is often used for logistic regression in which $\dot{y}\in\{-1,1\}$ a target and $z_l$ is output of a linear function $\smlmath{z_{l}=\sum_{k} W_{lk}x_{k}+b_{l}}$ where $W_{lk}$ is a weight from input $x_k$ to output $z_l$ and $b_l$ is bias for label $l$. A typical choice is, for instance, $w(\mathbf{y})=(|\mathbf{y}||\bar{\mathbf{y}}|)^{-1}$ as in BP-MLL.
In this work, we set $w(\mathbf{y})=1$, then the log loss above is equivalent to \emph{cross entropy} (CE), which is commonly used to train neural networks for classification tasks if we use \textit{sigmoid} transfer function in the output layer, i.e. $\smlmath{f_o(z) = 1/\left(1+\exp(-z)\right)}$, or simply $\smlmath{f_o(z)} =\sigma\left(z\right)$:
\begin{small}
\begin{equation}
J_{CE}(\Theta;\mathbf{x},\mathbf{y})=-\sum_{l}\left(y_{l}\log o_{l}) + (1-y_{l}) \log (1-o_{l})\right)
\label{eq:loss_ce}
\end{equation}
\end{small}
\noindent where $\smlmath{o_{l}}$ and $\smlmath{y_{l}}$ are the prediction and the target for label $l$, respectively.
Let us verify the equivalence between the \textit{log loss} and the \textit{CE}.
Consider the log loss function for only label $l$.
\begin{equation}
J_{log}(\Theta;\mathbf{x},y_{l})=\log(1+e^{-\dot{y}_{l}z_{l}})=-\log\left(\frac{1}{1+e^{-\dot{y}_{l}z_{l}}}\right)
\end{equation}
As noted,  $\dot{y}$  in the log loss takes either $-1$ or $1$, which allows us to split the above equation as follows:
\begin{equation}
-\log\left(\frac{1}{1+e^{-\dot{y}_{l}z_{l}}}\right) =
\left\{
\begin{array}{l l}
-\log\left(\sigma\left(z_{l}\right)\right) & \quad \textrm{if $\dot{y}=1$} \\
-\log\left(\sigma\left(-z_{l}\right)\right) & \quad \textrm{if $\dot{y}=-1$}
\end{array}
\right.
\end{equation}
\\Then, we have the corresponding CE by using a property of the sigmoid function
$
\sigma\left(-z\right)= 1-\sigma\left(z\right)
$
\begin{equation}
J_{CE}\left(\Theta;\mathbf{x},y_{l}\right)=-\left(y_{l}\log o_{l}+(1-y_{l})\log\left(1-o_{l}\right)\right)
\end{equation}
where $y\in\{0,1\}$ and $o_{l}=\sigma\left(z_l\right)$.

\paragraph*{\bf Computational Expenses} In addition to consistency with rank loss, CE has an advantage in terms of computational efficiency; computational cost for computing gradients of parameters with respect to PWE is getting more expensive as the number of labels grows.
The error term $\delta^{(2)}_{l}$ for label $l$ which is propagated
to the hidden layer is defined as
\begin{small}
\begin{equation}
\delta^{(2)}_{l} =
\left\{
  \begin{array}{l l}
  -\frac{1}{|\mathbf{y}||\bar{\mathbf{y}}|}{\displaystyle \sum_{n \in \bar{\mathbf{y}}}} \exp (-(o_{l} -o_{n}))f_{o}^{\prime}(z_{l}^{(2)}), \quad \text{if $l \in \mathbf{y}$ }\\
  \frac{1}{|\mathbf{y}||\bar{\mathbf{y}}|}{\displaystyle \sum_{p \in \mathbf{y}}} \exp(-(o_{p} - o_{l}))f_{o}^{\prime}(z_{l}^{(2)}), \quad \text{if $l \in \bar{\mathbf{y}}$}
  \end{array} \right.
\end{equation}
\end{small}
\noindent
Whereas the computation of
$\smlmath{\delta^{(2)}_{l}=-y_{l}/o_{l}+(1-y_{l})/(1-o_{l})f_{o}^{\prime}(z^{(2)}_{l})}$
for the CE can be performed efficiently, obtaining error terms $
\delta^{(2)}_{l} $ for the PWE is $L$ times more expensive than one in ordinary NN utilizing the cross entropy error function.
This also shows that BP-MLL scales poorly w.r.t. the number of unique labels.

\paragraph*{\bf Plateaus}
To get an idea of how differently both objective functions behave as a function of parameters to be optimized, let us draw graphs containing cost function values.
Note that it has been pointed out that the slope of the cost function as a function of the parameters plays an important role in learning parameters of neural networks \cite{Solla1988,Glorot2010} which we follow.

\begin{figure*}[t!]
    \centering
    \subfloat [Comparison of CE and PWE]{\label{fig:surf_nn_BP-MLL} \includegraphics[scale=0.3]{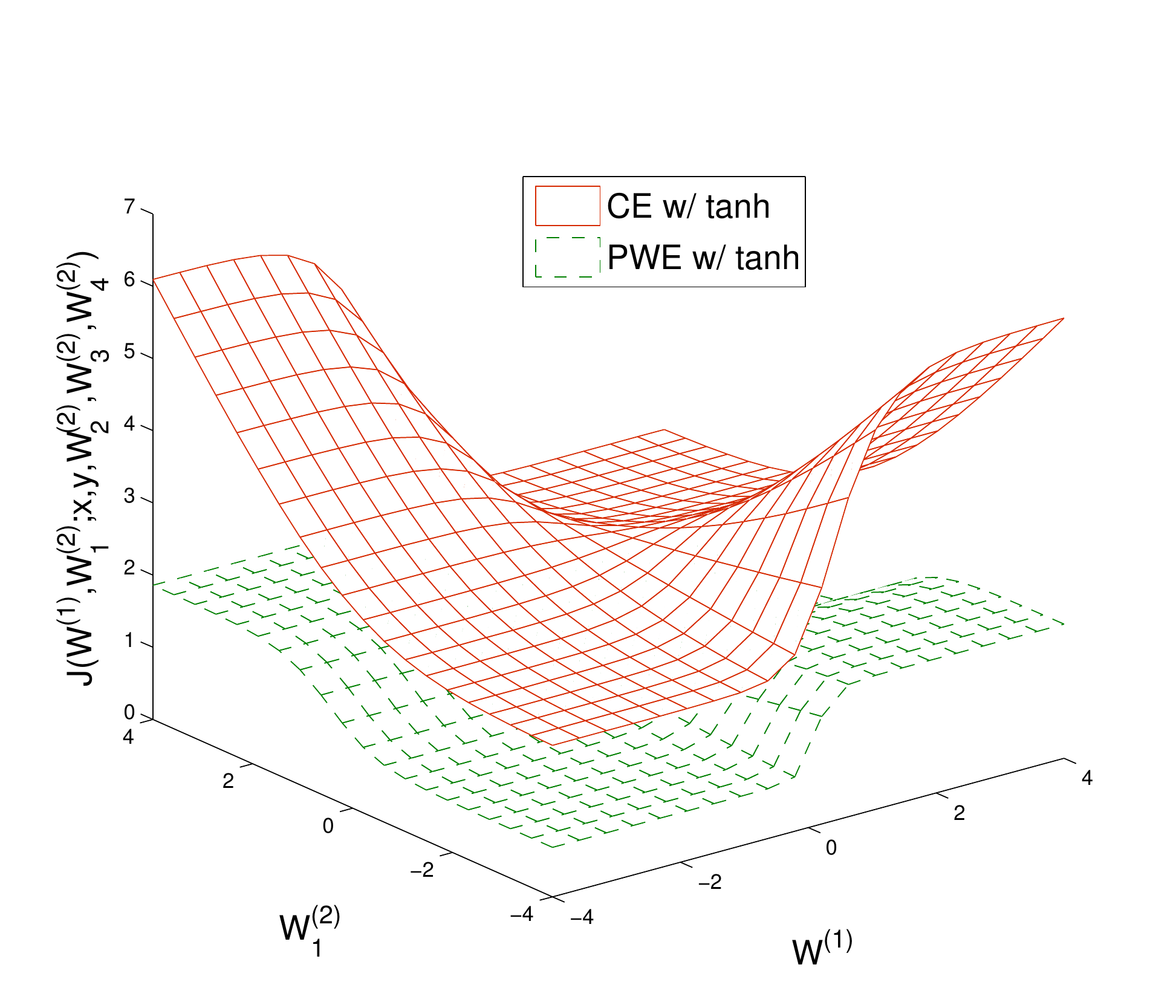}}
    \hfill
    \subfloat [Comparison of $\tanh$ and ReLU, both for CE]{\label{fig:surf_units_nn} \includegraphics[scale=0.3]{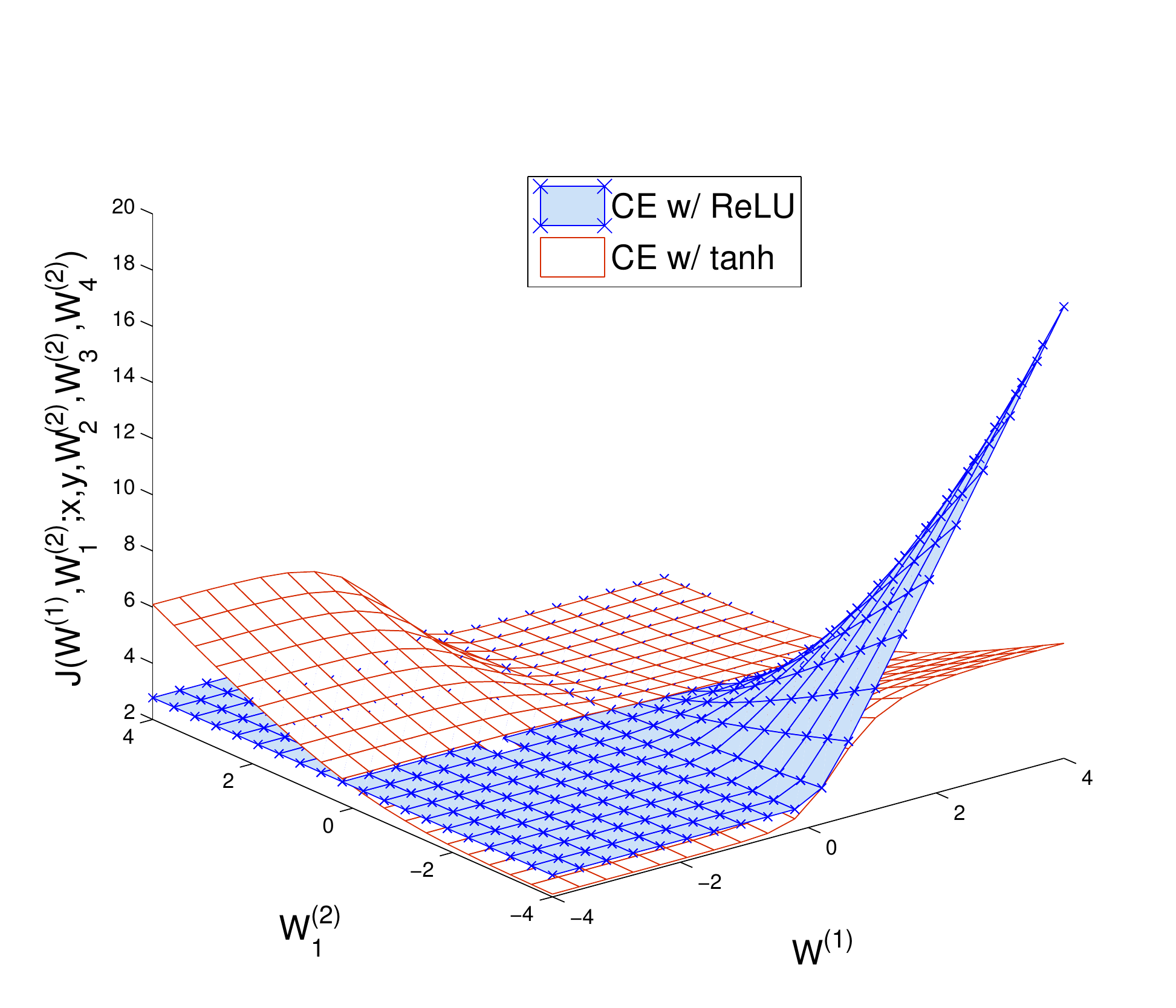}}
    \caption{
Landscape of cost functions and a type of hidden units.
$\smlmath{W^{(1)}}$ represents a weight connecting an input unit to a hidden unit. Likewise, $\smlmath{W^{(2)}_1}$ denotes a weight from the hidden unit to output unit 1. The \textit{z}-axis stands for a value for the cost function $\smlmath{J(W^{(1)},W^{(2)}_1;\mathbf{x},\mathbf{y},W^{(2)}_2,W^{(2)}_3,W^{(2)}_4)}$ where instances $\mathbf{x}$, targets $\mathbf{y}$ and weights $\smlmath{W^{(2)}_2,W^{(2)}_3,W^{(2)}_4}$ are fixed.
    }
\label{fig:opt_landscape}
\end{figure*}
Consider two-layer neural networks consisting of $\smlmath{W^{(1)}} \in
  \mathbb{R}$ for the first layer, $\smlmath{\mathbf{W}^{(2)}} \in
  \mathbb{R}^{4\times1}$ for the second, output layer. Since we are
interested in function values with respect to two parameters
$\smlmath{W^{(1)}}$ and $\smlmath{W^{(2)}_1}$ out of 5 parameters,
$\smlmath{\mathbf{W}^{(2)}_{\{2,3,4\}}}$ is set to a fixed value $c$. In this paper we use $c=0$.\footnote{
The shape of the functions is not changed even if we set $c$ to arbitrary value since it is drawn by function values in \textit{z}-axis with respect to only $\smlmath{W^{(1)}}$ and $\smlmath{W^{(2)}_1}$.
}
Figure \ref{fig:opt_landscape}~\protect\subref{fig:surf_nn_BP-MLL} shows different shapes of the functions
and slope steepness. 
In figure \ref{fig:opt_landscape}~\protect\subref{fig:surf_nn_BP-MLL} both curves have similar shapes, but the curve for PWE has plateaus in which gradient descent can be very slow in comparison with the CE.
Figure \ref{fig:opt_landscape}~\protect\subref{fig:surf_units_nn} shows that CE with ReLUs, which is explained the next Section, has a very steep slope compared to CE with tanh. Such a slope can accelerate convergence speed in learning parameters using gradient descent.
We conjecture that these properties might explain why our set-up
converges faster than the other configurations, and BP-MLL performs poorly in most cases in our experiments.

\subsection{Recent Advances in Deep Learning} \label{SEC_ADVANCES}

In recent neural network and deep learning literature, a number of
techniques were proposed to overcome the difficulty of learning neural
networks efficiently. In particular, we make use of ReLUs, AdaGrad, and Dropout training, which are briefly discussed in
the following.

\paragraph*{\bf Rectified Linear Units}

\emph{Rectified linear units} (ReLUs) have been proposed as activation units
on the hidden layer and shown to yield better generalization performance \cite{nair2010rectified,Glorot2011,zeiler2013rectified}.
A ReLU disables negative activation ($\text{ReLU}(x)=\max(0,x)$) so that the number of parameters to be learned decreases during the training.
This sparsity characteristic makes ReLUs advantageous over the
traditional activation units such as \textit{sigmoid} and
\textit{tanh} in terms of the generalization performance.

\paragraph*{\bf Learning Rate Adaptation with AdaGrad}

\emph{Stochastic gradient descent} (SGD) is a simple but effective technique for minimizing the objective functions of NNs (Equation \ref{eq:grad_update}).
When SGD is considered as an optimization tool, one of the problems is
the choice of the learning rate.
A common approach is to estimate the learning rate which gives lower training errors on subsamples of training examples \cite{lecun2012efficient} and then decrease it over time.
Furthermore, to accelerate learning speed of SGD, one can utilize momentum \cite{rumelhart1986learning}.

Instead of a fixed or scheduled learning rate, an \emph{adaptive learning rate} method, namely AdaGrad, was proposed \cite{Duchi2011}.
The method determines the learning rate at iteration $\tau$ by keeping previous gradients $\Delta_{1:\tau}$ to compute the learning rate for each dimension of parameters
$
\smlmath{\eta_{i,\tau} = \eta_{0}/\sqrt{\sum_{t=1}^{\tau} \Delta_{i,t}^{2}}}
$
where $i$ stands for an index of each dimension of parameters and $\eta_{0}$ is the initial learning rate and shared by all parameters.
For multi-label learning, it is often the case that a few labels occur
frequently, whereas the majority only occurs rarely, so that the rare
ones need to be updated with larger steps in the direction of the
gradient. If we use AdaGrad, the learning rates for the frequent
labels decreases because the gradient of the parameter for the
frequent labels will get smaller as the updates proceed. On the other
hand, the learning rates for rare labels remain comparatively large.

\paragraph*{\bf Regularization using Dropout Training}

In principle, as the number of hidden layers and hidden units in a
network increases, its expressive power also increases.
If one is given a large number of  training examples, training a larger networks will result in better performance than using a smaller one.
The problem when training such a large network is that the model is more prone to getting stuck in local minima due to the huge number of parameters to learn.
Dropout training \cite{hinton2012improving} is a technique for
preventing overfitting in a huge parameter space.
Its key idea is to decouple hidden units that activate the same output together,
by randomly dropping some hidden units' activations. Essentially, this
corresponds to training an ensemble of networks with smaller hidden
layers, and combining their predictions. However, the individual
predictions of all possible hidden layers need not be computed and
combined explicitly, but the output of the ensemble can be
approximately reconstructed from the full network. Thus, dropout
training has a similar regularization effect as ensemble techniques.

\section{Experimental Setup} \label{SEC_Experiment}
We have shown that why the structure of NNs needs to be reconsidered in the previous Sections.
In this Section, we describe evaluation measures to show how effectively NNs perform by combining recent development in learning neural networks based on the fact that the \textit{univariate} loss is consistent with respect to rank loss on large-scale textual datasets.

\paragraph*{\bf Evaluation Measures} \label{SEC_Measures}

Multi-label classifiers can be evaluated in two groups of measures: bipartition and ranking.
Bipartition measures operate on classification results, i.e. a set of labels assigned by classifiers to each document,
while ranking measures operate on the ranked list of labels.
In order to evaluate the quality of a ranked list, we consider several ranking measures \cite{Schapire2000}.
Given a document $\mathbf{x}$ and associated label information $\mathbf{y}$, consider a multi-label learner $f_{\theta}(\mathbf{x})$ that is able to produce scores for each label. These scores, then, can be sorted in descending order. Let $r(l)$ be the rank of a label $l$ in the sorted list of labels.
We already introduced \textit{Rank loss}, which is concerned primarily in this work, in Section \ref{SEC_DefRankLoss}.
\textit{One-Error} evaluates whether the top most ranked label with the highest score is is a positive label or not: $\mathbb{I}\left(r^{-1}{(1)}(f_{\theta}(x))\notin \mathbf{y}\right)$ where $r^{-1}(1)$ indicates the index of a label positioning on the first place in the sorted list. \textit{Coverage} measures on average how far one needs to go down the ranked list of labels to achieve recall of 100\%: $\max_{l_{i}\in \mathbf{y}} r(l_i)-1$. 
\textit{Average Precision} or AP measures the average fraction of labels preceding relevant labels in the ranked list of labels: $\frac{1}{|\mathbf{y}|}\sum_{l_{i} \in \mathbf{y}}\frac{|\{l_{j} \in \mathbf{y} | r(l_{j}) \leq r(l_{i})\}|}{r(l_{i})}$.

For bipartition measures, Precision, Recall, and $F_{1}$ score are conventional methods to evaluate effectiveness of information retrieval systems.
There are two ways of computing such performance measures: \textit{Micro-averaged} measures and \textit{Macro-averaged} measures\footnote{Note that scores computed by micro-averaged measures might be much higher than that by macro-averaged measures if there are many rarely-occurring labels for which the classification system does not perform well.
This is because macro-averaging weighs each label equally, whereas
micro-averaged measures are dominated by the results of frequent labels.}\cite{Manning2008}.
\vskip-10pt
\begin{scriptsize}
\[
P_{micro} = \frac{\sum_{l=1}^{L}tp_{l}}{\sum_{l=1}^{L}tp_{l}+fp_{l}}, R_{micro} = \frac{\sum_{l=1}^{L}tp_{l}}{\sum_{l=1}^{L}tp_{l}+fn_{l}}, F_{1-micro} = \frac{\sum_{l=1}^{L}2tp_{l}}{\sum_{l=1}^{L}2tp_{l}+fp_{l} + fn_{l}}
\]
\end{scriptsize}
\vskip-10pt
\begin{scriptsize}
\vskip-10pt
\[
P_{macro} = \frac{1}{L}\sum_{l=1}^{L}\frac{tp_{l}}{tp_{l}+fp_{l}}, R_{macro} = \frac{1}{L}\sum_{l=1}^{L}\frac{tp_{l}}{tp_{l}+fn_{l}}, F_{1-macro} = \frac{1}{L}\sum_{l=1}^{L}\frac{2tp_{l}}{2tp_{l}+fp_{l}+fn{l}}
\]
\end{scriptsize}
\paragraph*{\bf Datasets}
Our main interest is in large-scale text classification, for which we
selected six representative domains, whose characteristics are
summarized Table~\ref{Tab_stats}. 
For Reuters21578 we used the same training/test split as previous works \cite{Yang2012}.
Training and test data were switched for RCV1-v2 \cite{Lewis2004}
which originally consists of 23,149 train and 781,265 test documents.
The EUR-Lex, Delicious and Bookmarks datasets were taken from the
MULAN
repository.\footnote{\url{http://mulan.sourceforge.net/datasets.html}}
Except for Delicious and Bookmarks, all documents are represented with
\textit{tf-idf} features with cosine normalization such that length of the document vector is 1 in order to account for the different document lengths.

In addition to these standard benchmark datasets, we 
prepared a large-scale dataset from documents of the German Education Index (GEI).\footnote{\url{http://www.dipf.de/en/portals/portals-educational-information/german-education-index}} 
The GEI is a database of links to more than 800,000 scientific articles with metadata, e.g. title, authorship, language of an article and index terms.
We consider a subset of the dataset consisting of approximately 300,000 documents which have abstract as well as the metadata.
Each document has multiple index terms which are carefully hand-labeled by human experts with respect to the content of articles.
We processed plain text by removing stopwords and stemming each token.
To avoid the computational bottleneck from a large number of labels, we chose the 1,000 most common labels out of about 50,000.
We then randomly split the dataset into 90\% for training and 10\% for test.

\begin{table*}[t!]
\caption{Number of documents ($D$), size of vocabulary ($D$), total
  number of labels ($L$) and average number of labels per instance
  ($C$) for the six datasets used in our study.}
\setlength{\tabcolsep}{4pt}
\centering
\begin{footnotesize}
\begin{tabular}{c r r r r r}
\hline
Dataset                                     && $M$ & $D$ & $L$ & $C$ \\ \hline
Reuters-21578                       && 10789 & 18637 & 90 & 1.13 \\
RCV1-v2     && 804414 & 47236 & 103 & 3.24 \\
EUR-Lex                                     && 19348 & 5000 & 3993 & 5.31 \\
Delicious                                   && 16105 & 500 & 983 & 19.02 \\
Bookmarks                               && 87856 & 2150 & 208 & 2.03 \\
German Education Index          && 316061 & 20000 & 1000 & 7.16 \\
\hline
\end{tabular}
\end{footnotesize}
\label{Tab_stats}
\end{table*}

\paragraph*{\bf Algorithms}
\label{SEC_Alg}

Our main goal is to compare our NN-based approach to BP-MLL.
${\rm NN}_{\rm A} $ stands for the single hidden layer neural networks which have \textit{ReLUs} for its hidden layer and which are trained with SGD where each parameter of the neural networks has their own learning rate using \textit{AdaGrad}.
${\rm NN}_{\rm AD}$ additionally employs \textit{Dropout} based on the same settings as ${\rm NN}_{\rm A}$.
{\scriptsize T} and {\scriptsize R} following BP-MLL indicate \textit{tanh} and \textit{ReLU} as a transfer function in the hidden layer.
For both NN and BP-MLL, we used 1000 units in the hidden layer over all datasets. \footnote{The optimal number of hidden units of BP-MLL and NN was tested among 20, 50, 100, 500, 1000 and 4000 on validation datasets. Usually, the more units are in the hidden layer, the better performance of networks is. We chose it in terms of computational efficiency.}
As Dropout works well as a regularizer, no additional regularization to prevent overfitting was incorporated. 
The base learning rate $\eta_0$ was also determined among $\left[0.001, 0.01, 0.1\right]$ using validation data.

We also compared the NN-based algorithms to binary relevance (BR)
using SVMs (Liblinear) as a base learner, as a representative of the
state-of-the-art. 
The penalty parameter $C$ was optimized in the range of $[10^{-3},10^{-2}, \ldots , 10^2, 10^3]$ based on either average of micro- and macro-average F$_{1}$ or rankloss on validation set.
${\rm BR}_{\rm B}$ refers to linear SVMs where $C$ is optimized with bipartition measures on the validation dataset.
BR models whose penalty parameter is optimized on ranking measures are indicated as ${\rm BR}_{\rm R}$.
In addition, we apply the same thresholding technique which we utilize in our NN approach (Section \ref{SEC_Thresholding}) on a ranked list produced by BR models (${\rm BR}_{\rm R}$).

\section{Results}

We evaluate our proposed models and other baseline systems on datasets with varying statistics and characteristics.
We first show experimental results that confirm that  the techniques
discussed in Section~\ref{SEC_ADVANCES} actually contribute to an
increased performance of NN-based multi-label classification, and then
compare all algorithms on the six above-mentioned datasets in order to
get an overall impression of their performance.

\paragraph*{\bf Better Local Minima and Acceleration of Convergence Speed}

\begin{figure}[t]
\includegraphics[scale=0.32]{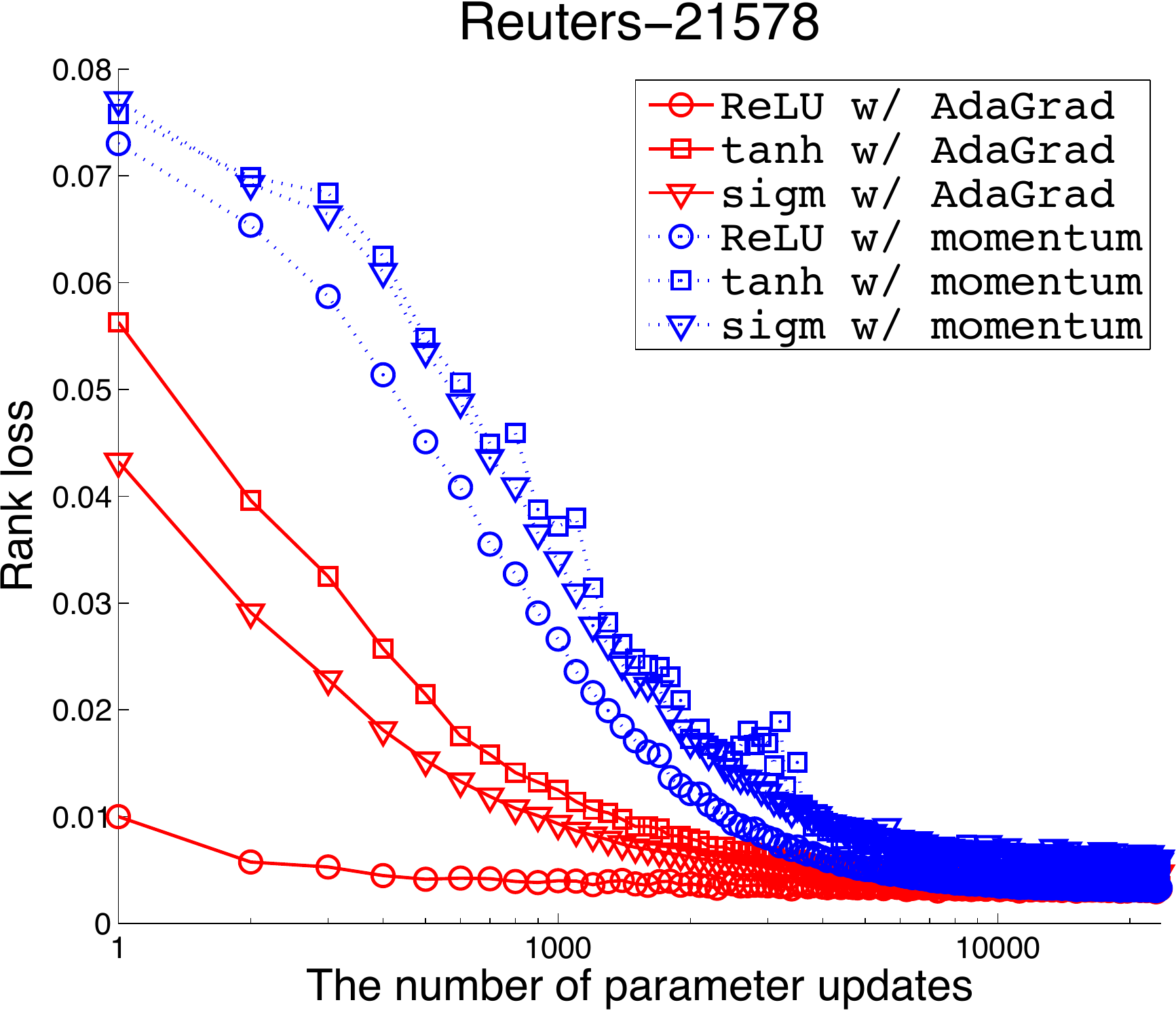}
\hfill
\includegraphics[scale=0.32]{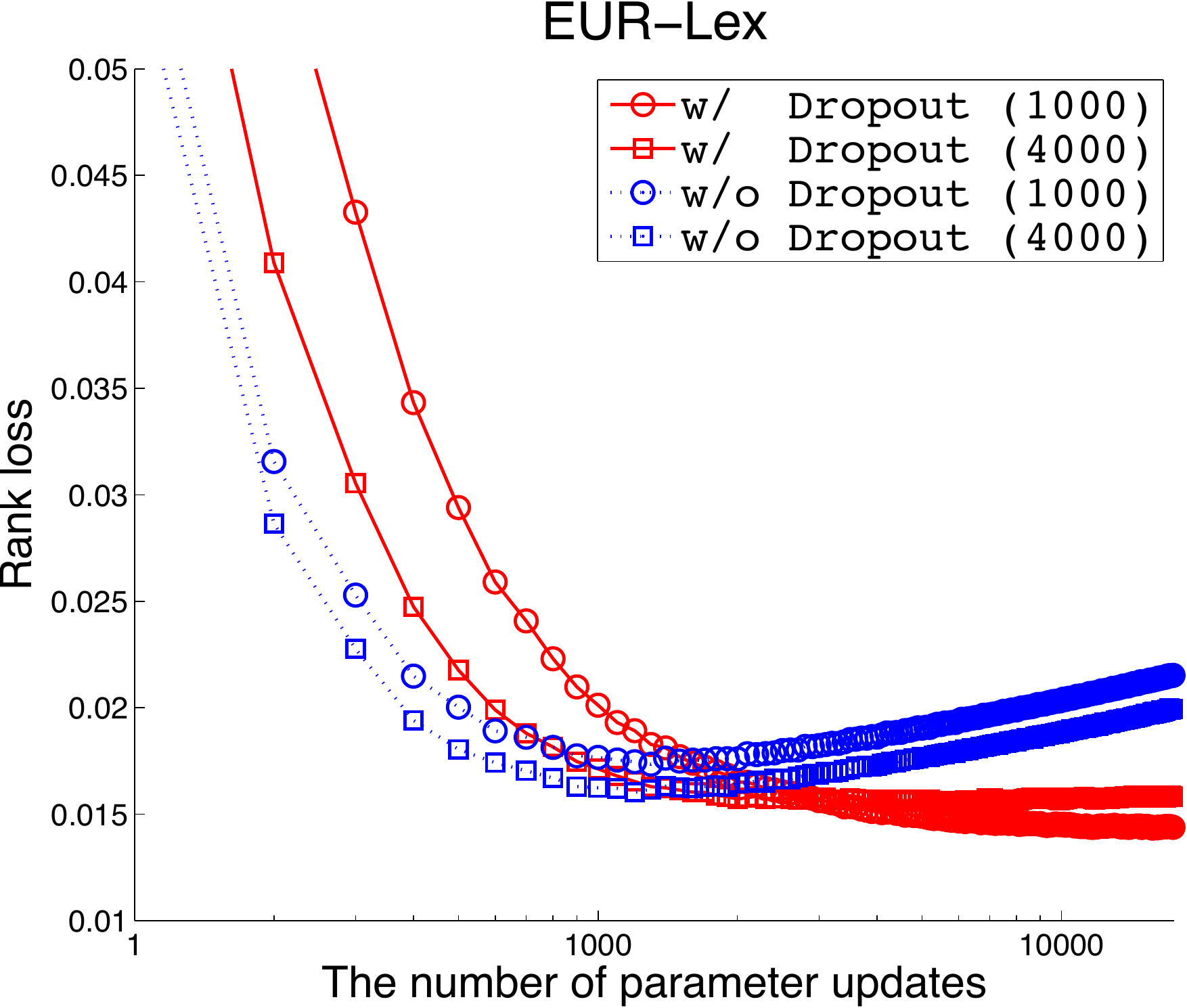}
\caption{
(\textit{left}) 
\label{fig:conv_speed}
effects of AdaGrad and momentum on three types of
transfer functions in the hidden layers in terms of rank loss on  Reuters-21578.
The number of parameter updates in \textit{x}-axis corresponds to the
number of evaluations of Eq. (\ref{eq:grad_update}).
(\textit{right}) \label{fig:drop_eurlex}
effects of dropout with two different numbers of hidden units in terms of rank loss on EUR-Lex.
}
\label{FIG_Effect}
\end{figure}
First we intend to show the effect of ReLUs and AdaGrad in terms of convergence speed and rank loss.
The left part of Figure~\ref{fig:conv_speed}  shows that all three
results of AdaGrad (red lines) show a lower rank loss than all three versions of
momentum. Moreover, within each group,  ReLUs outperform the versions
using $\tanh$ or sigmoid activation functions. That NNs with ReLUs 
at the hidden layer converge faster into a better weight space has
been previously observed for the speech domain
\cite{zeiler2013rectified}.\footnote{However, unlike the results of
  \cite{zeiler2013rectified}, in our preliminary experiments adding more hidden layers did
  not further improve generalization performance.}
This faster convergence is a 
major advantage of combining recently proposed learning components
such as ReLUs and AdaGrad, which facilitates a quicker learning of the
parameters of NNs. This is particularly important for the large-scale
text classification problems that are the main focus of this work.

\paragraph*{\bf Decorrelating Hidden Units While Output Units Remain Correlated}
One major goal of multi-label learners is to minimize rank loss by
leveraging inherent correlations in a label space. However, we
conjecture that these correlations also may cause overfitting because
if groups of hidden units specialize in predicting particular label
subsets that occur frequently in the training data, it will become
harder to predict novel label combinations that only occur in the test
set. Dropout effectively fights this by randomly dropping individual
hidden units, so that it becomes harder for groups of hidden units to
specialize in the prediction of particular output combinations, i.e.,
they decorrelate the hidden units, whereas the correlation of output units still remains.
Particularly, a subset of output activations $\mathbf{o}$ and hidden activations $\mathbf{h}$ would be correlated through $\mathbf{W}^{(2)}$.

We observed overfitting across all datasets except for Reuters-21578 and RCV1-v2 under our experimental settings.
The right part of Figure \ref{fig:drop_eurlex} shows how well Dropout prevents NNs from
overfitting on the test data of EUR-Lex.
In particular, we can see that with increasing numbers of parameter
updates, the performance of regular NNs eventually got worse in terms of rank loss.
On the other hand, when dropout is employed, 
convergence is initially slower, but eventually effectively prevents overfitting.

\paragraph*{\bf Limiting Small Learning Rates in BP-MLL}
\begin{figure}[t!]
\subfloat{\label{fig:rankloss_pwe_ce_lr} \includegraphics[scale=0.32]{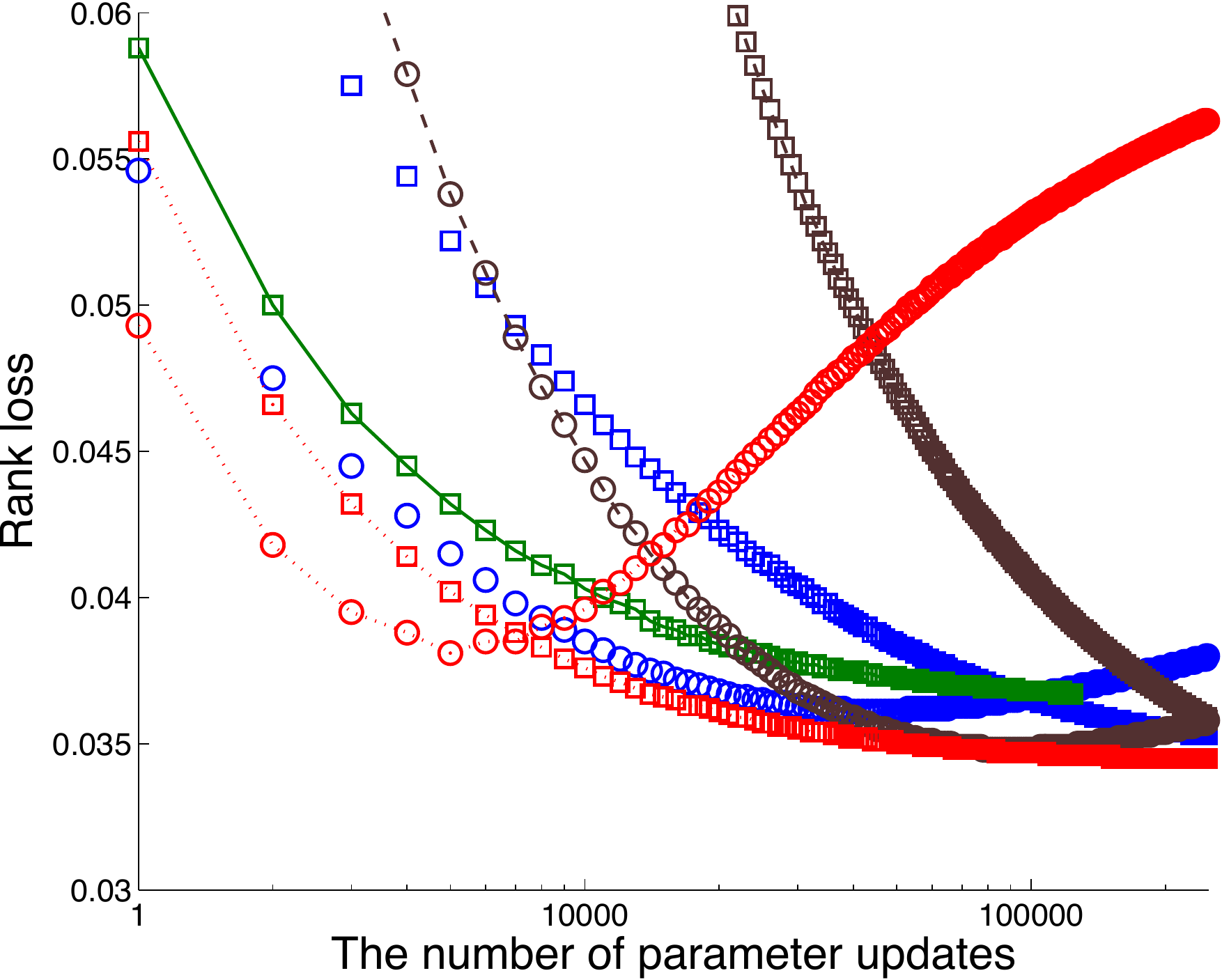}}
\hfill
\subfloat{\label{fig:map_pwe_ce_lr} \includegraphics[scale=0.32]{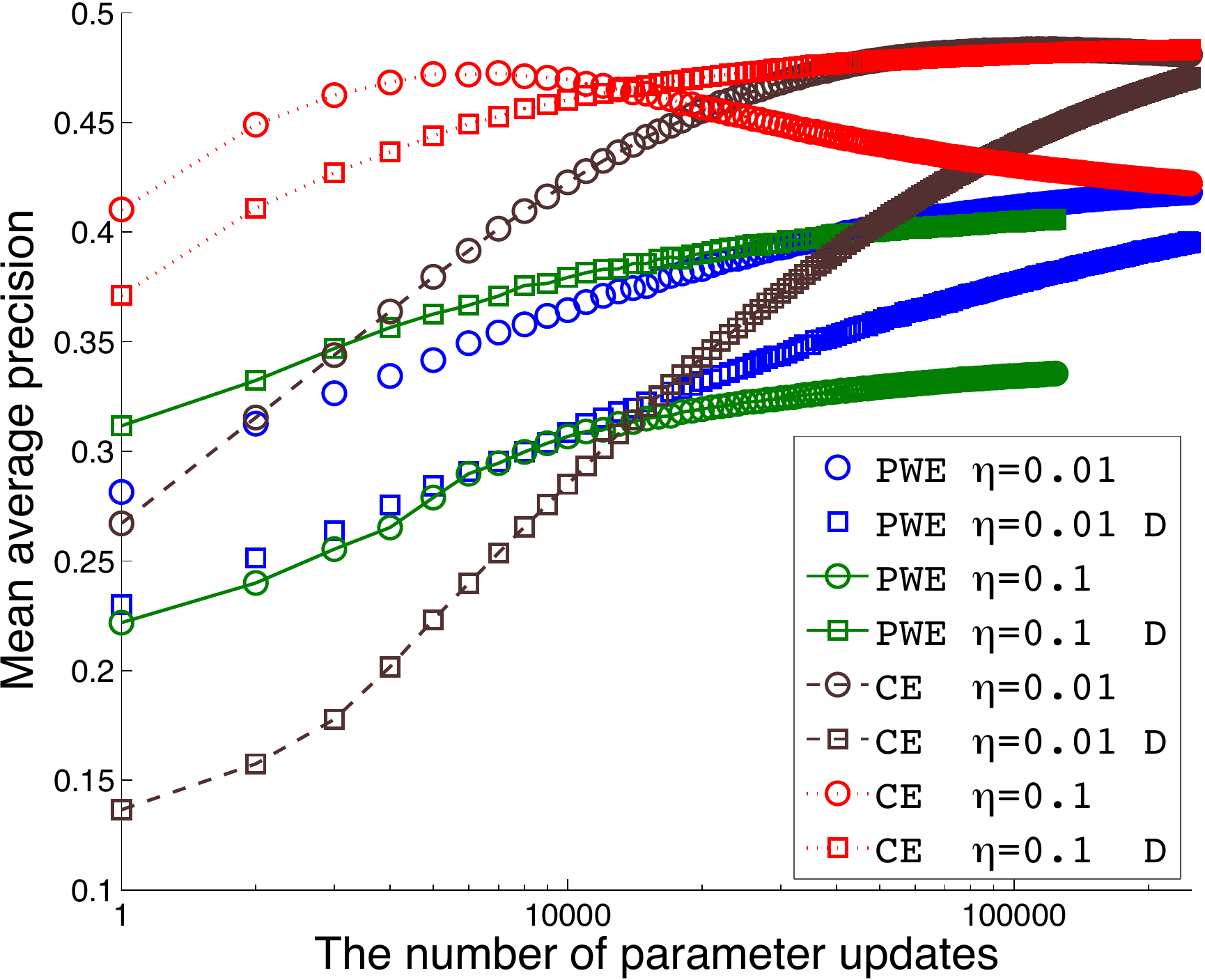}}
\caption{Rankloss (left) and mean average precision (right) on the German Education Index test data for the different cost functions.
$\eta$ denotes the base learning rate and D indicates that Dropout is applied. Note that \textit{x}-axis is in log scale.
}
\label{fig:learning_rates}
\end{figure}

The learning rate strongly influences convergence and learning speed \cite{lecun2012efficient}.
As we have already seen in the Figure \ref{fig:opt_landscape}, the
slope of PWE is less steep than CE, which implies that smaller
learning rates should be used.
Specifically, we observed PWE allows only smaller learning rate 0.01
(blue markers) in contrast with CE that works well a relatively larger
learning rate 0.1 (red markers) in Figure \ref{fig:learning_rates}.
In the case of PWE with the larger learning rate (green markers),
interestingly, dropout (rectangle markers in green) makes it converge
towards much better local minima, yet it is still worse than the other
configurations.  
It seems that the weights of BP-MLL oscillates in the vicinity of local minima and, indeed, converges \textit{worse} local minima.
However, it makes learning procedure of BP-MLL slow compared to NNs with CE making bigger steps for parameter updates.

With respect to Dropout, Figure \ref{fig:learning_rates} also shows that for the same learning rates, networks without Dropout converge much faster than ones working with Dropout in terms of both rank loss and MAP.
Regardless of the cost functions, overfitting arises over the networks without Dropout and it is likely that overfitting is avoided effectively as discussed earlier.\footnotetext{A trajectory for PWE $\eta=0.1$ is missing in the figure because it got 0.2 on the rankloss measure which is much worse than the other configurations.}

\paragraph*{\bf Comparison of Algorithms}

Table~\ref{Tab_ExpResults} shows detailed results of all experiments
with all algorithms on all six datasets, except that we could not
obtain results of BP-MLL on EUR-Lex within a reasonable time frame. In
an attempt to summarize the results, Table~\ref{Tab_ExpAvgRank} shows
the average rank of each algorithm in these six datasets according to
all ranking an bipartition measures discussed in Section~\ref{SEC_Measures}.

We can see that although BP-MLL focuses on minimizing pairwise ranking
errors,  
thereby capturing label dependencies, the single
hidden layer NNs with cross-entropy minimization (i.e., ${\rm NN}_{\rm
  A} $ and ${\rm NN}_{\rm AD} $) work much better not only on rank loss but also on other ranking measures.
The binary relevance (BR) approaches show acceptable performance on ranking measures even though label dependency was ignored during the training phase.
In addition, ${\rm NN}_{\rm A}$ and ${\rm NN}_{\rm AD}$ perform as
good as or better than other methods on bipartition measures as well
as on ranking measures.
\begin{table*}[t!]
\caption{Average ranks of the algorithms on ranking and bipartition
  measures. 
}
    \centering
    \begin{scriptsize}
	\begin{tabular}{| l | l | l | l | l || l | l | l | l | l | l |}
    \hline
    \multirow{2}{*} {Eval. measures} & \multicolumn{4}{ c ||}{Ranking} & \multicolumn{6}{ c |}{Bipartition} \\ \cline{2-11}
    & rankloss & oneError & Coverage & MAP & miP & miR & miF & maP & maR & maF \\ \hline
    \hline
    \multicolumn{11}{| c |}{\textbf{Average Ranks}} \\ \hline
    ${\rm NN}_{\rm A}$  &   2.2 &   2.4 &   2.6 &   2.2 &   \textbf{2}  &   6   &   2.4 &   \textbf{1.8}    &   5.6 &   \textbf{2}\\ \hline
    ${\rm NN}_{\rm AD}$     &   \textbf{1.2}    &   \textbf{1.4}    &   \textbf{1.2}    &   \textbf{1.6}    &   \textbf{2}  &   5.8 &   \textbf{1.8}    &   2   &   5.6 &   2.2\\ \hline
    ${\rm BP\text{-}MLL}_{\rm TA}$  &   5.9 &   6.9 &   6.2 &   6.2 &   6.6 &   3.6 &   6.6 &   6.8 &   3   &   6.2\\ \hline
    ${\rm BP\text{-}MLL}_{\rm TAD}$     &   5   &   5.7 &   5   &   5.7 &   7   &   3.6 &   7   &   7   &   4   &   7.2\\ \hline
    ${\rm BP\text{-}MLL}_{\rm RA}$  &   5.2 &   7   &   5.4 &   6.6 &   5.6 &   \textbf{2.8}    &   5   &   5   &   \textbf{2.8}    &   4.2\\ \hline
    ${\rm BP\text{-}MLL}_{\rm RAD}$     &   3.1 &   6.3 &   3   &   5.8 &   6   &   \textbf{2.8}    &   5.8 &   5.6 &   3.6 &   5.6\\ \hline
    ${\rm BR}_{\rm B}$  &   7.4 &   3.3 &   6.9 &   4.3 &   3.2 &   6.8 &   4.6 &   4.4 &   6.8 &   5.6\\ \hline
    ${\rm BR}_{\rm R}$  &   6   &   3   &   5.7 &   3.6 &   3.6 &   4.6 &   2.8 &   3.4 &   4.6 &   3\\ \hline
\end{tabular}
\end{scriptsize}
\label{Tab_ExpAvgRank}
\end{table*}

We did not observe significant improvements by replacing hidden units of BP-MLL from tanh to ReLU.
However, if we change the cost function in the previous setup from PWE to CE, significant improvements were obtained.
Because ${\rm BP\text{-}MLL}_{\rm RAD}$ is the same architecture as
${\rm NN}_{\rm AD}$ except for its cost function,\footnote{For PWE we use \textit{tanh} in the output layer, but \textit{sigmoid} is used for CE because predictions $\mathbf{o}$ for computing CE with targets $\mathbf{y}$ needs to be between 0 and 1.}
we can say that the differences in the effectiveness of NNs and BP-MLL
are due to the use of different cost functions.
This also implies that the main source of improvements for NNs against BP-MLL is replacement of the cost function.
Again, Figure \ref{fig:learning_rates} shows the difference between two cost functions more explicitly.

\begin{table*}[p]
    \caption{
Results on ranking and bipartition measures.
Results for BP-MLL on EUR-Lex are missing because the runs could not
be completed in a reasonably short time.}
    \centering
\begin{scriptsize}
    \begin{tabular}{| l | l | l | l | l || l | l | l | l | l | l |}
    \hline
    \multirow{2}{*} {Eval. measures} & \multicolumn{4}{ c ||}{Ranking} & \multicolumn{6}{ c |}{Bipartition} \\ \cline{2-11}
    & rankloss & oneError & Coverage & MAP & miP & miR & miF & maP & maR & maF \\ \hline
    \hline
    \multicolumn{11}{| c |}{\textbf{Reuters-21578}} \\ \hline
    ${\rm NN}_{\rm A}$ & 0.0037 & 0.0706 & 0.7473 & 0.9484 & 0.8986 & 0.8357 & 0.8660 & \textbf{0.6439} & 0.4424 & 0.4996  \\ \hline
    ${\rm NN}_{\rm AD}$ & \textbf{0.0031} & 0.0689 & \textbf{0.6611} & 0.9499 & 0.9042 & 0.8344 & 0.8679 & 0.6150 & 0.4420 & 0.4956 \\ \hline
    ${\rm BP\text{-}MLL}_{\rm TA}$ & 0.0039 & 0.0868 & 0.8238 & 0.9400 & 0.7876 & 0.8616 & 0.8230 & 0.5609 & \textbf{0.4761} & 0.4939 \\ \hline
    ${\rm BP\text{-}MLL}_{\rm TAD}$ & 0.0039 & 0.0808 & 0.8119 & 0.9434 & 0.7945 & 0.8654 & 0.8284 & 0.5459 & 0.4685 & 0.4831 \\ \hline    
    ${\rm BP\text{-}MLL}_{\rm RA}$ & 0.0054 & 0.0808 & 1.0987 & 0.9431 & 0.8205 & 0.8582 & 0.8389 & 0.5303 & 0.4364 & 0.4624 \\ \hline
    ${\rm BP\text{-}MLL}_{\rm RAD}$ & 0.0063 & 0.0719 & 1.2037 & 0.9476 & 0.8421 & 0.8416 & 0.8418 & 0.5510 & 0.4292 & 0.4629 \\ \hline
    ${\rm BR}_{\rm B}$  & 0.0040 & \textbf{0.0613} & 0.8092 & \textbf{0.9550} & \textbf{0.9300} & 0.8096 & 0.8656 & 0.6050 & 0.3806 & 0.4455 \\  \hline
    ${\rm BR}_{\rm R}$ & 0.0040 & \textbf{0.0613} & 0.8092 & \textbf{0.9550} & 0.8982 & \textbf{0.8603} & \textbf{0.8789} & 0.6396 & 0.4744 & \textbf{0.5213} \\     \hline
    \hline
    \multicolumn{11}{| c |}{\textbf{RCV1-v2}} \\ \hline
    ${\rm NN}_{\rm A}$ & 0.0040 & 0.0218 & 3.1564 & 0.9491 & 0.9017 & 0.7836 & 0.8385 & 0.7671 & 0.5760 & 0.6457 \\ \hline
    ${\rm NN}_{\rm AD}$ & \textbf{0.0038} & \textbf{0.0212} & \textbf{3.1108} & \textbf{0.9500} & \textbf{0.9075} & 0.7813 & 0.8397 & \textbf{0.7842} & 0.5626 & 0.6404 \\ \hline
    ${\rm BP\text{-}MLL}_{\rm TA}$ & 0.0058 & 0.0349 & 3.7570 & 0.9373 & 0.6685 & 0.7695 & 0.7154 & 0.4385 & 0.5803 & 0.4855 \\ \hline
    ${\rm BP\text{-}MLL}_{\rm TAD}$ & 0.0057 & 0.0332 & 3.6917 & 0.9375 & 0.6347 & 0.7497 & 0.6874 & 0.3961 & 0.5676 & 0.4483 \\ \hline
    ${\rm BP\text{-}MLL}_{\rm RA}$ & 0.0058 & 0.0393 & 3.6730 & 0.9330 & 0.7712 & 0.8074 & 0.7889 & 0.5741 & 0.6007 & 0.5823 \\ \hline
    ${\rm BP\text{-}MLL}_{\rm RAD}$ & 0.0056 & 0.0378 & 3.6032 & 0.9345 & 0.7612 & 0.8016 & 0.7809 & 0.5755 & 0.5748 & 0.5694 \\ \hline
    ${\rm BR}_{\rm B}$ & 0.0061 & 0.0301 & 3.8073 & 0.9375 & 0.8857 & 0.8232 & \textbf{0.8533} & 0.7654 & 0.6342 & 0.6842 \\ \hline
    ${\rm BR}_{\rm R}$ & 0.0051 & 0.0287 & 3.4998 & 0.9420 & 0.8156 & \textbf{0.8822} & 0.8476 & 0.6961 & \textbf{0.7112} & \textbf{0.6923} \\ \hline
    \hline
    \multicolumn{11}{| c |}{\textbf{EUR-Lex}} \\ \hline
    ${\rm NN}_{\rm A}$ & 0.0195 & 0.2016 & 310.6202 & 0.5975 & 0.6346 & 0.4722 & 0.5415 &  0.3847 & 0.3115 & 0.3256 \\ \hline
    ${\rm NN}_{\rm AD}$ & \textbf{0.0164} & \textbf{0.1681} & \textbf{269.4534} & \textbf{0.6433} & \textbf{0.7124} & 0.4823 & \textbf{0.5752} &  \textbf{0.4470} & 0.3427 & 0.3687 \\ \hline
    ${\rm BR}_{\rm B}$ & 0.0642 & 0.1918 & 976.2550 & 0.6114 & 0.6124 & 0.4945 & 0.5471 & 0.4260 & \textbf{0.3643} & \textbf{0.3752} \\ \hline
    ${\rm BR}_{\rm R}$ & 0.0204 & 0.2088 & 334.6172 & 0.5922 & 0.0329 & \textbf{0.5134} & 0.0619 & 0.2323 & 0.3063 & 0.2331 \\ \hline
    \hline
    \multicolumn{11}{| c |}{\textbf{German Education Index}} \\ \hline
    ${\rm NN}_{\rm A}$ & \textbf{0.0350} & 0.2968 & 138.5423 & \textbf{0.4828} & 0.4499 & 0.4200 & \textbf{0.4345} & 0.4110 & 0.3132 & \textbf{0.3427}  \\ \hline
    ${\rm NN}_{\rm AD}$ & 0.0352 & \textbf{0.2963} & 138.3590 & 0.4797 & 0.4155 & 0.4472 & 0.4308 & 0.3822 & 0.3216 & 0.3305 \\ \hline
    ${\rm BP\text{-}MLL}_{\rm TA}$ & 0.0386 & 0.8309 & 150.8065 & 0.3432 & 0.1502 & \textbf{0.6758} & 0.2458 & 0.1507 & \textbf{0.5562} & 0.2229 \\ \hline
    ${\rm BP\text{-}MLL}_{\rm TAD}$ & 0.0371 & 0.7591 & 139.1062 & 0.3281 & 0.1192 & 0.5056 & 0.1930 & 0.1079 & 0.4276 & 0.1632 \\ \hline
    ${\rm BP\text{-}MLL}_{\rm RA}$ & 0.0369 & 0.4221 & 143.4541 & 0.4133 & 0.2618 & 0.4909 & 0.3415 & 0.3032 & 0.3425 & 0.2878 \\ \hline
    ${\rm BP\text{-}MLL}_{\rm RAD}$ & 0.0353 & 0.4522 & \textbf{135.1398} & 0.3953 & 0.2400 & 0.5026 & 0.3248 & 0.2793 & 0.3520 & 0.2767 \\ \hline
    ${\rm BR}_{\rm B}$ & 0.0572 & 0.3052 & 221.0968 & 0.4533 & \textbf{0.5141} & 0.2318 & 0.3195 & 0.3913 & 0.1716 & 0.2319\\ \hline
    ${\rm BR}_{\rm R}$ & 0.0434 & 0.3021 & 176.6349 & 0.4755 & 0.4421 & 0.3997 & 0.4199 & \textbf{0.4361} & 0.2706 & 0.3097 \\ \hline
    \hline
    \multicolumn{11}{| c |}{\textbf{Delicious}} \\ \hline
    ${\rm NN}_{\rm A}$ & 0.0860 & 0.3149 & 396.4659 & 0.4015 & \textbf{0.3637} & 0.4099 & 0.3854 & 0.2488 & 0.1721 & 0.1772  \\ \hline
    ${\rm NN}_{\rm AD}$ & \textbf{0.0836} & \textbf{0.3127} & \textbf{389.9422} & \textbf{0.4075} & 0.3617 & 0.4399 & \textbf{0.3970} & \textbf{0.2821} & 0.1777 & \textbf{0.1824} \\ \hline
    ${\rm BP\text{-}MLL}_{\rm TA}$ & 0.0953 & 0.4967 & 434.8601 & 0.3288 & 0.1829 & 0.5857 & 0.2787 & 0.1220 & 0.2728 & 0.1572 \\ \hline
    ${\rm BP\text{-}MLL}_{\rm TAD}$ & 0.0898 & 0.4358 & 418.3618 & 0.3359 & 0.1874 & 0.5884 & 0.2806 & 0.1315 & 0.2427 & 0.1518 \\ \hline
    ${\rm BP\text{-}MLL}_{\rm RA}$ & 0.0964 & 0.6157 & 427.0468 & 0.2793 & 0.2070 & \textbf{0.5894} & 0.3064 & 0.1479 & \textbf{0.2609} & 0.1699 \\ \hline
    ${\rm BP\text{-}MLL}_{\rm RAD}$ & 0.0894 & 0.6060 & 411.5633 & 0.2854 & 0.2113 & 0.5495 & 0.3052 & 0.1650 & 0.2245 & 0.1567 \\ \hline
    ${\rm BR}_{\rm B}$ & 0.1184 & 0.4355 & 496.7444 & 0.3371 & 0.1752 & 0.2692 & 0.2123 & 0.0749 & 0.1336 & 0.0901 \\ \hline
    ${\rm BR}_{\rm R}$ & 0.1184 & 0.4358 & 496.8180 & 0.3371 & 0.2559 & 0.3561 & 0.2978 & 0.1000 & 0.1485 & 0.1152 \\ \hline
    \hline
    \multicolumn{11}{| c |}{\textbf{Bookmarks}} \\ \hline
    ${\rm NN}_{\rm A}$ & 0.0663 & 0.4924 & 22.1183 & 0.5323 & 0.3919 & 0.3907 & 0.3913 & 0.3564 & 0.3069 & 0.3149 \\ \hline
    ${\rm NN}_{\rm AD}$ & \textbf{0.0629} & \textbf{0.4828} & \textbf{20.9938} & \textbf{0.5423} & \textbf{0.3929} & 0.3996 & \textbf{0.3962} & \textbf{0.3664} & 0.3149 & \textbf{0.3222} \\ \hline
    ${\rm BP\text{-}MLL}_{\rm TA}$ & 0.0684 & 0.5598 & 23.0362 & 0.4922 & 0.0943 & 0.5682 & 0.1617 & 0.1115 & 0.4743 & 0.1677 \\ \hline
    ${\rm BP\text{-}MLL}_{\rm TAD}$ & 0.0647 & 0.5574 & 21.7949 & 0.4911 & 0.0775 & \textbf{0.6096} & 0.1375 & 0.0874 & \textbf{0.5144} & 0.1414 \\ \hline
    ${\rm BP\text{-}MLL}_{\rm RA}$ & 0.0707 & 0.5428 & 23.6088 & 0.5049 & 0.1153 & 0.5389 & 0.1899 & 0.1235 & 0.4373 & 0.1808 \\ \hline
    ${\rm BP\text{-}MLL}_{\rm RAD}$ & 0.0638 & 0.5322 & 21.5108 & 0.5131 & 0.0938 & 0.5779 & 0.1615 & 0.1061 & 0.4785 & 0.1631 \\ \hline
    ${\rm BR}_{\rm B}$ & 0.0913 & 0.5318 & 29.6537 & 0.4868& 0.2821 & 0.2546 & 0.2676 & 0.1950 & 0.1880 & 0.1877\\ \hline
    ${\rm BR}_{\rm R}$ & 0.0895 & 0.5305 & 28.7233 & 0.4889 & 0.2525 & 0.4049 & 0.3110 & 0.2259 & 0.3126 & 0.2569\\ \hline
    \end{tabular}
\end{scriptsize}
    \label{Tab_ExpResults}
\end{table*}

\section{Conclusion} \label{SEC_Conclusion}

This paper presents a multi-label classification framework based on a
neural network and a simple threshold label predictor.
We found that our approach outperforms BP-MLL, both in predictive
performance as well as in computational complexity and convergence speed.
We have explored why BP-MLL as a multi-label text classifier does not
perform well.
Our experimental results showed the proposed framework is an effective method for the multi-label text classification task.
Also, we have conducted extensive analysis to characterize the
effectiveness of combining ReLUs with AdaGrad for fast convergence
rate, and utilizing Dropout to prevent overfitting which results in
better generalization.

\medskip
\begin{small}
\paragraph{\textbf{Acknowledgments}}
This work has been supported by the Information Center for Education
of the German Institute for Educational Research (DIPF) under the
Knowledge Discovery in Scientific Literature (KDSL) program.
\end{small}

\bibliographystyle{splncsnat}

\bibliography{mlc}

\end{document}